\documentclass[10pt,a4paper]{article}
\usepackage[utf8x]{inputenc}
\usepackage{ucs}
\usepackage{amsmath}
\usepackage{amsfonts}
\usepackage{amssymb}
\usepackage{graphicx}
\usepackage[left=3.00cm, right=3.00cm, top=2.50cm, bottom=2.50cm]{geometry}

\usepackage{bm}

\usepackage{tikz}
\usetikzlibrary{trees}
\usetikzlibrary{shadows}
\usetikzlibrary{backgrounds}
\usetikzlibrary{shapes,arrows,automata}
\usetikzlibrary{positioning,intersections,fit,calc}

\usetikzlibrary{bayesnet}

\usepackage{hl_short}

\usepackage{listings}
\usepackage{xcolor}
\usepackage[justification=centering]{caption}

\definecolor{mygreen}{rgb}{0,0.6,0}
\definecolor{mygray}{rgb}{0.5,0.5,0.5}
\definecolor{mymauve}{rgb}{0.58,0,0.82}
\definecolor{stringcolor}{RGB}{0,128,0}
\definecolor{commentcolor}{RGB}{128,128,128}
\definecolor{cleargray}{RGB}{240,240,240}
\definecolor{blueamidst}{HTML}{7a94cc}
\definecolor{kwcolor}{rgb}{0.13,0.13,1}
\lstdefinestyle{terminal}{
	numbers=none
}

\lstdefinestyle{xml}{
	language=xml, 
	morekeywords={repositories, repository,id,url,dependencies,dependency,groupId,artifactId,scope,},
}

\lstdefinestyle{java}{
	language=java, 
	basicstyle=\fontfamily{lmvtt}\selectfont\small,
	numberstyle=\color{black},
	commentstyle=\color{commentcolor},
	keywordstyle=\color{kwcolor}\bfseries,
	stringstyle=\color{stringcolor},
	numbersep=3pt,
	numberstyle=\tiny\color{mygray},
}

\newcommand{\code}{Code Fragment}

\renewcommand{\newpage}{}

\newcommand{\pkg}[1]{#1}
\newcommand{\proglang}[1]{#1}
\newcommand{\citep}[1]{\cite{#1}}

\newcommand{\linebreakGen}{\linebreak} 

\title{AMIDST: a Java Toolbox for Scalable Probabilistic Machine Learning}

\author{
	Andr\'{e}s R. Masegosa\thanks{These four authors are considered as first authors and contributed equally to this work.} 
	\\
	The Norwegian University\\ of Science and Technology
	\and 
	Ana M. Mart\'{\i}nez$^\star$
	\\
	Aalborg University
	\and
	Dar\'{i}o Ramos-L\'{o}pez$^\star$
	\\
	University of Almer\'{\i}a
	\and
	Rafael Caba\~{n}as$^\star$
	\\
	Aalborg University
	\and 
	Antonio Salmer\'{o}n
	\\
	University of Almer\'{\i}a
	\and
	Thomas D. Nielsen
	\\
	Aalborg University
	\and
	Helge Langseth
	\\
	The Norwegian University\\ of Science and Technology
	\and
	Anders L. Madsen
	\\
	Aalborg University\\ HUGIN EXPERT A/S
}

\date{}

\begin{document}
	
	\maketitle

	\begin{abstract}
The \pkg{AMIDST} Toolbox is a software for scalable probabilistic machine learning with a special focus on (massive) streaming data. The toolbox supports a flexible modeling language based on probabilistic graphical models with latent variables and temporal dependencies. The specified models can be learnt
from large data sets using parallel or distributed implementations of Bayesian learning
algorithms for either streaming or batch data. These algorithms are based on a flexible variational message passing scheme, which supports
discrete and continuous variables from a wide range of probability distributions. \pkg{AMIDST} also leverages existing functionality and algorithms by interfacing to software tools such as Flink, Spark, MOA, Weka, R and HUGIN. \pkg{AMIDST} is an open source toolbox written in Java and available at \url{http://www.amidsttoolbox.com} under the Apache Software License version 2.0.
	\end{abstract}

\vspace{2mm}


\section{Introduction}\label{sec:introduction}


\pkg{AMIDST}   is a toolbox for the analysis of small and large-scale data sets using probabilistic machine learning methods based on graphical models. A probabilistic graphical model (PGM) is a framework consisting of two parts: a qualitative component in the form of a graphical model encoding conditional independence assertions about the domain being modeled; a quantitative component consisting of a collection of local  probability distributions adhering to the independence properties specified in the graphical model. Collectively, the two components provide a compact representation of the joint probability distribution over the set of variables in the domain being modelled. We consider the Bayesian network (BN) framework \citep{Pearl88} and its dynamic extension in the form of 2-Timeslice BNs (2TBNs) \citep{murphy2002dynamic}. These frameworks provide a well-founded and principled approach for performing inference in complex (temporal) domains endowed with uncertainty.\newline  

\pkg{AMIDST}  implements parallel and distributed algorithms 
for learning 
probabilistic graphical models with latent or unobserved variables such as Gaussian
mixtures, (probabilistic) principal component analysis, hidden Markov
models, (switching) linear dynamical systems, latent Dirichlet allocation, etc. More generally, the toolbox is
 able to learn any user-defined probabilistic (graphical) model belonging to the conjugate exponential family 
using novel message passing algorithms \citep{masegosa2016d}. Additionally, the \pkg{AMIDST}  Toolbox has the possibility of handling large data streams
by implementing (approximate) Bayesian updating. To the best of our knowledge, there is no existing software for mining data streams based on a wide range of PGMs (including latent variable models); most existing tools focus on stationary data sets \citep{murphy2007software} (see Figure~\ref{fig:Taxonomy}).\newline


\begin{figure}[h!bt]
\begin{center}
\includegraphics[width=0.55\linewidth]{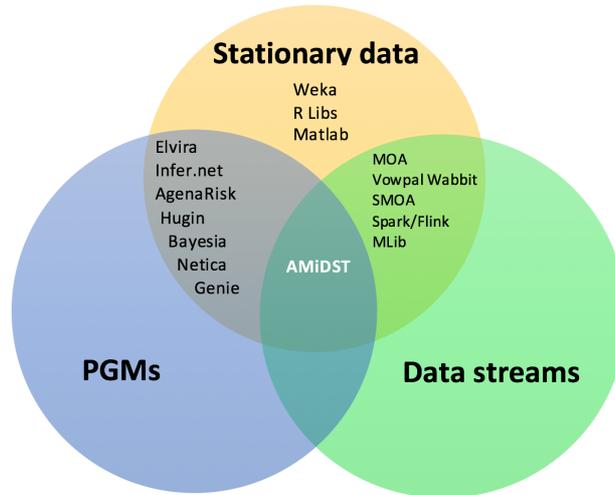}
%
%
%
%
%
%
%
%
%


\caption{\label{fig:Taxonomy} A taxonomy of data mining software.}
\end{center}
\end{figure}

In this way, \pkg{AMIDST}'s approach to machine learning is based on the use of openbox models that can be inspected and which can incorporate prior information or knowledge about the domain, in contrast to other approaches based on blackbox models, which cannot be interpreted by the users. This is why the focus of \pkg{AMIDST} is not only to learn models for making predictions rather than to learn models to extract knowledge from the data \citep{borchani2015modeling, borchani2015dynamic}. And, as commented above, with the advantage that this toolbox is equipped with a general learning engine implementing scalable variational Bayesian inference algorithms, which allow to learn models exploiting a wide range of computational settings. The use of a Bayesian approach not only allows us to deal with model uncertainty problems but also naturally deals with data streams through the standard operation of Bayesian updating (see Section \ref{sec:BatchStreaming}). Moreover, the probabilistic approach we pursue gives a basis to naturally deal with the presence of missing data, which is a quite ubiquitous problem in real-world applications. 

\begin{figure}[h!bt]
\begin{center}
\includegraphics[width=0.70\linewidth]{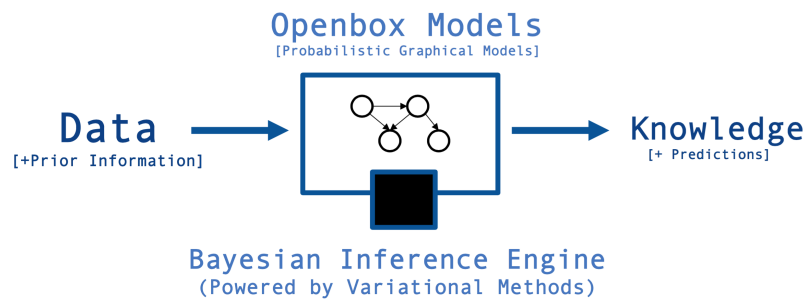}
\caption{\label{fig:Taxonomy} AMIDST's approach to machine learning.}
\end{center}
\end{figure}

The toolbox has been developed by the {AMIDST}  Consortium\footnote{http://www.amidst.eu}, which includes both academic and industrial partners.   Our software has been used in the automotive industry \citep{weidl2015early} and in the finance sector \citep{borchani2015dynamic,borchani2015modeling}. The \pkg{AMIDST}  Toolbox can be used, for instance, for classification, clustering, regression, and density estimation tasks \citep{murphy2012machine}. \pkg{AMIDST}  also offers the ability to create, learn and perform inference in (dynamic) BNs.





\tikzset{
  pics/my plateau/.style args={#1}{
      code = {

  \node[obs]          (X)   {$\bmX$}; %
  
  \node[latent, above= 0.6 of X]          (H)   {$\bmH$}; %
  \edge{H}{X};
  
  \node[latent, above= 0.6  of H]  (theta)   {$\bmtheta_{#1}$}; %
  \edge{theta}{H};\draw[bend right,->]  (theta) to (X);

  \plate {Observations} { %
    (X)(H)
  } {\tiny $i={1,\ldots, N}$} ;
  
  \node[const, above left = 1 and 1.2 of X]  (slave)   {Slave #1}; %
  
  \node [below= 0.1 of slave](computer){\includegraphics[scale=0.07]{figs/computer-29686.pdf}};
  
  \draw [dashed,rounded corners=10pt](-2.8,-1.2) rectangle (8mm,32mm);
   
   }}
}

\section{The {AMIDST} Toolbox features}\label{sec:toolbox}
In what follows, we describe the main features of the \pkg{AMIDST} Toolbox. 


\subsection{Probabilistic graphical models}
\label{section:PGMs}

The \pkg{AMIDST} Toolbox supports the specification of \textit{Bayesian networks (BNs)} \citep{Pearl88,JensenNielsen2007}, which  are widely used PGMs for reasoning under uncertainty. They graphically encode a set of conditional independence assumptions that can be exploited to efficiently perform a wide variety of inference tasks such as marginal belief computation, belief update, abductive inference, etc.  Formally, let $\bm{X} = \{X_1,\ldots,X_N\}$ denote the set of  random variables defining our domain problem. BNs can be graphically represented by a directed acyclic graph (DAG). Each node, labelled $X_i$ in the graph, is associated with a factor or conditional probability $p(X_i|Pa(X_i))$, where $Pa(X_i)\subset \bm X\setminus X_i$ represents the so-called \emph{parent variables} of $X_i$. Additionally, for each parent $X_j \in Pa(X_i)$, the graph contains one directed edge pointing from $X_j$ to the \emph{child} variable $X_i$. A BN defines a joint distribution $p(\bm X)$ in the following form:

\vspace{-4mm}
\begin{equation}
p(\bm X) = \prod_{i=1}^N p(X_i|Pa(X_i))
\end{equation}

%

Traditionally, BNs have been defined for discrete domains, where the entities of interest are modelled by discrete variables which ensures that belief updating can be performed efficiently and in closed form \citep{Pearl88}. However, this requirement imposes severe restrictions as many domains contain entities that are more appropriately modelled by variables with continuous state spaces. To deal with this problem, the \pkg{AMIDST} Toolbox allows the specification of  conditional linear gaussian Networks (CLG) \citep{Lauritzen1992,lauritzen1996graphical}, which are hybrid BNs\footnote{The toolbox is designed to be extended to support  any model belonging to the conjugate-exponential family (this class includes many of the most common distributions, such as Gaussian, mixture of Gaussian, Exponential, Dirichlet, Bernoulli, Gamma, etc.). }.  In the CLG model, the conditional distribution of each discrete variable $X_D\in \bmX$ given its parents is a multinomial, whilst the conditional distribution of each continuous variable $Z\in\bmX$ with discrete parents $\bmX_D\subseteq\bmX$  and continuous parents $\bmX_C\subseteq\bmX$, is given by

\begin{equation}\label{eq:contprob}
p(z | \bmX_D = \bmx_D, \bmX_C = \bmx_C) = \mathcal{N}(z; \alpha(\bmx_D) + \beta(\bmx_D)^\intercal \bmx_C, \sigma(\bmx_D) ),
\end{equation}

\noindent for all $\bmx_D \in \Omega_{\bmX_D}$ and $\bmx_C \in \Omega_{\bmX_C}$,where $\alpha$ and $\beta$ are the coefficients of a linear regression model of $Z$ given its continuous parents; this model can differ for each configuration of the discrete variables $\bmX_D$.\newline



%
%
%
%

\begin{figure}
	\centering
	\resizebox{0.20\linewidth}{!}{
	\begin{tikzpicture}
	
	
	\node[obs]          (X)   {$\bmX_i$}; %
	
	\node[latent, right= 1.5 of X]          (H)   {$\bmH_i$}; %
	\edge{H}{X};
	
	\node[latent, above= 1  of H]  (theta)   {$\bmtheta$}; %
	\edge{theta}{H};\edge{theta}{X};
	
	\node[const, left= 0.5  of theta]  (alpha)   {$\bmalpha$}; %
	\edge{alpha}{theta};
	
	\plate {Observations} { %
		(X)(H)
	} {\tiny $i={1,\ldots, N}$} ;

	\end{tikzpicture}
}
	\caption{\label{fig:plateau} An example of a PGM covered by \pkg{AMIDST} (where $N$ is the number of samples, $\bmX_i$ a set of features, $\bmH_i$ a set of local latent variables, $\bmtheta$ the global parameters and $\bmalpha$ its hyper-parameters).}
\end{figure}
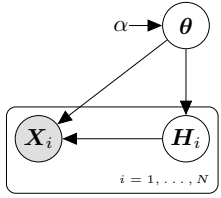

Furthermore, latent (i.e., hidden) variables are supported.  These variables cannot be observed and are included in the model to capture correlation structure. The use of latent variables allows the representation of a large range of problems with complex probabilistic dependencies. Supported PGMs generally fit to the plate structure \citep{Buntine1994operations} depicted in Figure \ref{fig:plateau}, although more general PGMs can be considered. \newline

The \pkg{AMIDST} Toolbox also allows the specification of \textit{dynamic BNs (DBNs)}, which are used to model domains that evolve over time by representing explicitly the temporal dynamics of the system. In the DBN framework, variables are indexed  by a discrete time index $t$. For example, $X_t$ is the variable $X$ at time $t$. In this way, we explicitly model the state of the system at any given time. For reasoning over time, we need to model the joint probability $p(\bm X_{1:T})$ which has the following natural cascade decomposition:

$$p(\bm X_{1:T})  = \prod_{t=1}^T p(\bm X_t|\bm X_{1:t-1}),$$

\noindent where $p(\bm X_t|\bm X_{1:t-1})$ is equal to $p(\bm X_1)$ for $t=1$, and $\bm X_{1:t-1}$ is the set  $\{X_1,X_2,\ldots X_{t-1}\}$.
\subsection{Scalable Bayesian inference}
The \pkg{AMIDST} Toolbox offers Bayesian parameter learning functionality using powerful approximate and scalable algorithms based on 
variational methods. \pkg{AMIDST} relies on the \textit{variational message passing} (VMP) algorithm \citep{WinnBishop2005}. Two versions are provided; a parallel version exploiting multi-core architectures, powered by \proglang{Java} 8 Streams \citep{masegosa2016probabilistic};  and a novel distributed version, named d-VMP \citep{masegosa2016d}, for large-scale data processing on computing clusters running  either Apache Flink\footnote{\url{http://flink.apache.org}} or Apache Spark\footnote{\url{http://spark.apache.org}}. The d-VMP method has been used in the context of financial data analysis to perform efficient scalable inference in a model with more than one billion nodes (see \citep{masegosa2016d} for details). Stochastic Variational Inference \citep{SVIHoffmanEtAl2013} is also available for learning.\newline 

 In relation to standard inference, we also provide a parallel implementation of the \textit{importance sampling} algorithm \citep{hammersley1964monte,CAEPIA2015}.  Versions of these methods for dynamic models are supported by means of the Factored Frontier algorithm \citep{FactoredFrontier}. Scalable abductive inference \citep{ramos2016scalable} is also supported using optimization algorithms and Monte Carlo methods. 

%

\subsection{Batch-streaming learning}
\label{sec:BatchStreaming}
The toolbox has the possibility to handle massive data sets and large data streams. Massive data sets can be processed in distributed computer clusters by interfacing with Apache Flink and Spark using novel scalable and distributed variational methods \citep{masegosa2016d}. We use a Bayesian approach for updating the model as new data arrives. Assuming $\bmX_t$ is a new batch of data at time $t$, $\bmtheta$ the parameters of the model and $\bmH$ the set of latent variables, the Bayesian updating is performed as follows,
\begin{equation}\label{eq:batchupdate}
p(\bmtheta,\bmH|\bmX_1,\ldots,\bmX_t) \propto p(\bmX_t|\bmtheta,\bmH) p(\bmtheta,\bmH | \bmX_1,\ldots,\bmX_{t-1}).
\end{equation}

\noindent This provides a  natural setting for stream processing, and  there is therefore no need to learn the model from scratch as new data arrives. When learning from streaming data \pkg{AMIDST} also provides an implementation of  the \textit{streaming variational Bayes} method \citep{broderick2013streaming}, also based on VMP and d-VMP algorithms. \newline


An important aspect of streaming data is that the domain
being modeled is often \emph{non-stationary}. That is, the distribution governing the data changes over time.
This situation, which is known as \textit{concept drift} \citep{Gama2014}, can be detected with \pkg{AMIDST} using a novel probabilistic approach \citep{borchani2015modeling}. 



\vspace{-1mm}
\subsection{Modular design}
The \pkg{AMIDST} Toolbox has been designed following a modular structure, i.e., all the provided functionality is
organized in different modules (Table \ref{tab:modules} shows a brief description of all the modules in the toolbox).
This modularity allows future extensions to be made independently
of the core design, thereby leaving the kernel small and robust. Another added value of the modularity is that 
it enables a more seamlessly interaction with external software. Currently, \pkg{AMIDST} interfaces with MOA\footnote{\url{http://moa.cms.waikato.ac.nz}}, Weka\footnote{\url{http://www.cs.waikato.ac.nz/ml/weka/}},  and HUGIN\footnote{\url{http://www.hugin.com}}.\newline

\begin{table}[h!bt]
	\centering
\begin{tabular}{|p{40mm}||p{90mm}|}
	\hline
	\textbf{Module}&\textbf{Description}	\\\hline\hline
	core
	&	
	\textit{Functionality related to (static) BNs  (i.e., learning and inference algorithms, classes handling data streams, static DAGs, variables, etc.).}
	\\\hline
	core-dynamic
	&
	\textit{Analogous to the core module, but implementing the functionality related to DBNs.}
	\\\hline
	latent-variable-models
	&
\textit{Contains a wide range of predefined (static and dynamic) latent variable models. Learning and applying these models is straight-forward, which simplifies the use of the toolbox.}
	\\\hline	
		lda
	&
	\textit{Allows text processing by means of the latent Dirichlet allocation model.}
	\\\hline	
	huginlink
	&
	\textit{Implements the interaction with the external software HUGIN (e.g., use of its inference engine, reading and writing BNs in HUGIN format, etc.)  }
	\\\hline	
	wekalink	
	&
\textit{	Implements the interaction with the external software Weka. For example, for creating a wrapper for evaluating an \pkg{AMDIST} classifier with Weka.}

	\\\hline	
	flinklink	
	&
\textit{	Implements the  integration with Apache Flink, which allows  the implemented algorithms to run in a distributed architecture.}
	\\\hline	
	sparklink	
	&
\textit{	Implements the  integration with Apache Spark which allows  the implemented algorithms to run in a distributed architecture.}	
	\\\hline
	moalink	
	&
		Implements the integration with MOA library. Any model developed on AMIDST can be evaluated within MOA.
	\\\hline	
			examples
	&
	\textit{A  set of code examples using the functionality provided in the rest of the modules. The unit tests are integrated int each module separately.}
	\\\hline			
\end{tabular}
\caption{Modules in \pkg{AMIDST}.}\label{tab:modules}
\end{table}

	\section{Code-examples}
\subsection{Managing data streams in a single computer}

The \texttt{DataStream} class in package \texttt{eu.amidst.core.datastream} is an interface for dealing with data streams stored on a single computer. The whole \pkg{AMIDST} Toolbox is specifically designed to process the data sequentially without having to load all observations into main memory simultaneously. In this way, this class can also handle static data-sets which do not fit into main memory. A \texttt{DataStream} object is composed by a collection of \texttt{DataInstance} objects. The functionality for loading data is provided by class \texttt{eu.amidst.core.io.DataStreamLoader}. In particular, an example of reading data from a .arff file is given in \code~\ref{code:readingStatic}.\newline 

\begin{lstlisting}[
caption={Reading a (static) data stream from a .arff file.}, 
label=code:readingStatic,
language=java, 
]
DataStream<DataInstance> data = DataStreamLoader.open("data0.arff");

\end{lstlisting}

When loading dynamic data streams (i.e., with temporal information), we will use the class \texttt{eu.amidst.dynamic.io.DataStreamLoader} instead. Now the returned object data stream will be a collection of \texttt{DynamicDataInstance} objects as shown in \code~\ref{code:redingDynamic}.\newline


\begin{lstlisting}[
caption={Loading a dynamic data stream from a .arff file.}, 
label=code:redingDynamic,
style=java, 
]
DataStream<DynamicDataInstance> data = DynamicDataStreamLoader.open("data0.arff");

\end{lstlisting}

Static and dynamic data is managed exactly in the same way as they are both handled with the interface \texttt{DataStream}. In \code~\ref{code:managingData} we show how to access to the attributes' names and types (lines 2 to 6), to print all the instances (line 9), and to save the data in a file (line 12). Note that, for accessing  each instance in the data stream, \textit{lambda expressions} provided by \proglang{Java} 8 are used.\newline

\begin{lstlisting}[
caption={Processing data in a multi-core environment.}, 
label=code:managingData,
style=java, 
]
// Print the attributes names and types
data.getAttributes().forEach(att -> {
	String name = att.getName();
	StateSpaceTypeEnum type = att.getStateSpaceType().getStateSpaceTypeEnum();
	System.out.println(name +" "+type.name());
});

// Print all the instances
data.stream().forEach(dataInstance -> System.out.println(dataInstance));

// Save the data stream into a file 
DataStreamWriter.writeDataToFile(data, "data0_copy.arff");
\end{lstlisting}

\newpage 
In line 9 of \code~\ref{code:managingData}, all the instances are sequentially processed. To do the processing  in parallel, we will use the method \texttt{parallelStream(int batchSize)} instead of \texttt{stream()}. Internally, data instances are grouped into batches and all the instances in the same batch are processed with the same thread.  The first lines of the output generated from the previous code are shown below.\newline  

\begin{lstlisting}[
caption={Output generated by \code~\ref{code:managingData}. This data-set contains two special attributes specific for dynamic data: \texttt{SEQUENCE\_ID} and \texttt{TIME\_ID}.}, 
label=code:outputManagingData,
style=terminal, 
]
SEQUENCE_ID REAL
TIME_ID REAL
DiscreteVar0 FINITE_SET
GaussianVar0 REAL
GaussianVar1 REAL
GaussianVar2 REAL
GaussianVar3 REAL
GaussianVar4 REAL
GaussianVar5 REAL
GaussianVar6 REAL
GaussianVar7 REAL
GaussianVar8 REAL
GaussianVar9 REAL
{SEQUENCE_ID = 0.0, TIME_ID = 0.0, DiscreteVar0 = 1.0, GaussianVar0 = 7.89477533117581, GaussianVar1 = 1.0534927992122323, GaussianVar2 = 8.002205739682081, GaussianVar3 = 4.438782227521086, GaussianVar4 = 0.4743537026415879, GaussianVar5 = -4.360599142940955, GaussianVar6 = -2.3040164637846487, GaussianVar7 = 1.479131531697834, GaussianVar8 = 6.418327748167779, GaussianVar9 = 4.021688685889985, }
{SEQUENCE_ID = 0.0, TIME_ID = 1.0, DiscreteVar0 = 1.0, GaussianVar0 = -6.474414081999592, GaussianVar1 = 7.040927942549324, GaussianVar2 = -19.893603802179552, GaussianVar3 = 3.0344777889027923, GaussianVar4 = 3.3646294832430335, GaussianVar5 = 15.65274347488313, GaussianVar6 = -4.156649241840171, GaussianVar7 = -19.851052439447308, GaussianVar8 = -20.560256855030985, GaussianVar9 = 27.166372803488336, }
. . .
\end{lstlisting}


\subsection{Managing data in a distributed computing cluster}

When dealing with large scale data, it might be convenient to have it stored in a distributed file system. In \pkg{AMIDST}, distributed data can be managed using Flink (module \textit{flinklink}). \code~\ref{code:flinkLoading} shows how to load data from a \textit{Hadoop Distributed File System (HDFS)}. In line 2, the Flink session is set up, and in line 5 the data is loaded using the static method \texttt{open} from  class \texttt{eu.amidst.flinklink.core.io.DataFlinkLoader}. This method returns an object of class \texttt{DataFlink<DataInstance>} and takes three arguments: the Flink execution enviroment, the path to the data, and a boolean indicating whether the data should be normalized or not.\newline 

\newpage 
\begin{lstlisting}[
caption={Loading distributed data in AMIDST.}, 
label=code:flinkLoading,
style=java, 
]
// Set-up Flink session
final ExecutionEnvironment env = ExecutionEnvironment.getExecutionEnvironment();

// Load the distributed data
DataFlink<DataInstance> data = DataFlinkLoader.open(env, "hdfs://dataFlink0.arff", false);
\end{lstlisting}

The access to the instances is slightly different compared to the multi-core \linebreak case (\code~\ref{code:managingData}). As shown in \code~\ref{code:distributedManaging}, in a distributed cluster, we \linebreak will use the method \texttt{getDataSet()} from  class \texttt{DataFlink<DataInstance>}, \linebreak which returns an object of class \texttt{DataSet} (i.e., a class from the Flink API). The writing operation is done by invoking the static method \texttt{writeDataToARFFFolder} from the class \texttt{eu.amidst.flinklink} \texttt{.core.io.DataFlinkWriter}.\newline

\begin{lstlisting}[
caption={Processing data in a distribtued environment.}, 
label=code:distributedManaging,
style=java, 
]
// Print all the instances from the distributed file
data.getDataSet().collect().forEach(dataInstance -> System.out.println(dataInstance));

// Save the data into a distributed ARFF folder
DataFlinkWriter.writeDataToARFFFolder(data, "hdfs://dataFlink0_copy.arff"); 
\end{lstlisting}

\subsection{Scalable learning of probabilistic graphical models}

\subsubsection{Learning a model}

The \pkg{AMIDST} Toolbox contains a wide range of predefined models, most of them including latent variables (see Table \ref{tab:latent-var-models}). These models are available in the module \textit{latent-variable-models}. All of them are direct instantations of the class of models supported (see Section \ref{section:PGMs}).\newline 

\newcommand{\widthCell}{50mm}	
\begin{table}[h!bt] 
	\centering
	\resizebox{0.90\linewidth}{!}{
		\begin{tabular}{|c|c|}
			\hline Static&Dynamic\\\hline 

			\begin{minipage}[t]{\widthCell}
											\vspace{0.5mm}
				\begin{itemize}\setlength\itemsep{0.001mm}
					\item Na\"ive Bayes (NB)
					\item (G)AODE/HODE					
					\item Gaussian Discriminant Analysis
					\item Gaussian Mixture
					\item Multivariate Gaussian Distribution
				\end{itemize}
							\vspace{0.5mm}
			\end{minipage}				\begin{minipage}[t]{\widthCell}
										\vspace{0.5mm}
			\begin{itemize}\setlength\itemsep{0.001mm}
				\item Latent Classification Models (LCM)
				\item Bayesian Linear Regression
				\item Factor Analysis (FA)
				\item Mixture of FA
			\end{itemize}
		\end{minipage}		&
			\begin{minipage}[t]{\widthCell}
											\vspace{0.5mm}
				\begin{itemize}\setlength\itemsep{0.001mm}
									\item Dynamic NB
					\item  Dynamic LCM
					\item Hidden Markov Model (HMM)
					\item (Switching) Linear Dynamical Systems
				\end{itemize}
			\end{minipage}
						\begin{minipage}[t]{\widthCell}
														\vspace{0.5mm}
							\begin{itemize}\setlength\itemsep{0.001mm}
								\item Factorial HMM
								\item Auto-regressive HMM
								\item Input-Output HMM
							\end{itemize}
						\end{minipage}						
			\\\hline
		\end{tabular}
	}
	\caption{Predefined models in AMIDST.}
	\label{tab:latent-var-models}
\end{table}

Learning and applying these models is straight-forward. As an example,  \code~\ref{code:StaticModelLearning}  shows how to learn a static   a \textit{Gaussian mixture} model \citep{murphy2012machine} with  a binary  latent variable  and continuous observable variables. The class \texttt{eu.amidst.} \texttt{latentvariablemodels.staticmodels.Model} encapsulates all the functionality for learning the predefined static models and hence, any class implementing a particular static model should inherit it. In lines 2 to 4, the model is built from a list of attributes. In line 6, by invoking the method \texttt{updateModel}, the probability distributions are learnt from data.  This method takes one argument, which can be either an object of class \texttt{DataStream<DataInstance>}  or \texttt{DataFlink<DataInstance>}, which corresponds to a local multi-core computer or a distributed computing cluster, respectively.  Note that, in this way, the code for learning a model is independent of the processing environment (i.e., multi-core or distributed). Finally, an object of class \texttt{BayesianNetwork} is obtained and printed to the standard output (lines 7 and 8).  An additional advantage of using \pkg{AMIDST} is that the code is flexible: if we aim to learn another predefined static model, we simply have to change the constructor used in line 3.\newline

\begin{lstlisting}[
caption={Learning a predefined static model from data.}, 
label=code:StaticModelLearning,
style=java, 
]
// Build the model
Model model =
			new GaussianMixture(data.getAttributes())
			.setNumStatesHiddenVar(2);

model.updateModel(data);  // Learn the distributions
BayesianNetwork bn = model.getModel();  // Obtain the learnt BN
System.out.println(bn);  // Print the BN

\end{lstlisting}

The output generated by \code~\ref{code:StaticModelLearning}  is shown below. The print-out includes the distributions associated to each variable:  the discrete variables have a multinomial, while  the continuous ones have  normal distributions given by \eqref{eq:contprob}. Note that, for space restriction, the generated output has been reduced.\newline 


\vspace{-2mm}

\begin{lstlisting}[
caption={Reduced output generated by \code~\ref{code:StaticModelLearning}.}, 
label=code:outputStaticModelLearnin,
style=terminal, 
]
Bayesian Network:
P(GaussianVar0 | HiddenVar) follows a Normal|Multinomial
Normal [ mu = 15.438615720911814, var = 2.4076525429420337 ] | {HiddenVar = 0}
Normal [ mu = 15.826209029022205, var = 2.596445695611393 ] | {HiddenVar = 1}

P(GaussianVar1 | HiddenVar) follows a Normal|Multinomial
Normal [ mu = -1.0677878081725678, var = 0.3767293765940319 ] | {HiddenVar = 0}
Normal [ mu = -1.8720749210971157, var = 0.36008303248004486 ] | {HiddenVar = 1}

. . .

P(GaussianVar9 | HiddenVar) follows a Normal|Multinomial
Normal [ mu = -1.4559663235292983, var = 1.6416477403801821 ] | {HiddenVar = 0}
Normal [ mu = -2.839310405569177, var = 1.440090565070485 ] | {HiddenVar = 1}

P(HiddenVar) follows a Multinomial
[ 0.543445287416122, 0.456554712583878 ]

\end{lstlisting}

\subsubsection{Updating a model}

An advantage of the \pkg{AMIDST} Toolbox is that the model can be easily updated when new data becomes available (see Equation~\eqref{eq:batchupdate}). This capacity is useful when analysing streaming data, as new data  is constantly being generated and it could happen that the old model might not accurately represent this new data (see \code~\ref{code:updatingModel}).\newline

\vspace{-4mm}

\begin{lstlisting}[
caption={Updating the probability distributions of a model with new data.}, 
label=code:updatingModel,
style=java, 
]
//Update the model with new information
for(int i=1; i<12; i++) {
	data = DataStreamLoader.open(path+"data"+i+".arff");
	model.updateModel(data);
	System.out.println(model.getModel());
}
\end{lstlisting}

\subsubsection{Learning models with temporal dependencies}

  When learning a dynamic model, an object of any class that inherits from \linebreak\texttt{eu.amidst.latentvariablemodels.dynamicmodels.DynamicModel} should be created. As an example, \code~\ref{code:learningDynamicModel} shows how a \textit{Kalman Filter}  can be learnt from dynamic data (i.e., an object of class \texttt{DataStream<DynamicDataInstance>}).

\begin{lstlisting}[
caption={Learning a predefined dynamic model from data.}, 
label=code:learningDynamicModel,
style=java, 
]
//Learn the model
DynamicModel model =
				new KalmanFilter(data.getAttributes())
				.setNumHidden(2);

model.updateModel(data); //Learn the distributions
DynamicBayesianNetwork dbn = model.getModel(); //Obtain the learnt dynamic BN
System.out.println(dbn); // Print the dynamic BN and save it
\end{lstlisting}

\subsubsection{Define custom models}

If neither of the predefined models fit with the problem being modelled, a custom model can be defined. For doing that, we should create a class that inherits \texttt{Model} and overrides  the method \texttt{protected void buildDAG()}. This class must contain a constructor taking as argument an object of class  \texttt{Attributes}. \code~\ref{code:customModel} shows the definition of a Gaussian mixture model with a local latent parent of each observed variable as an example of this procedure.  The most important part of the code is the method \texttt{buildDAG()}, where the member variable \texttt{dag} is initialized. An object of class \texttt{DAG} is basically a list of parent sets, one for each variable. Thus, to define a DAG, we should add the corresponding variables to these parent sets (lines 20 and 21).

\newpage 

\begin{lstlisting}[
caption={Definition of a custom model.}, 
label=code:customModel,
style=java, 
]
public class CustomModel extends Model{
public CustomModel(Attributes attributes) throws WrongConfigurationException {
	super(attributes);
}
@Override
protected void buildDAG() {
	//Obtain the observed variables
	List<Variable> attrVars = vars.getListOfVariables().stream().collect(Collectors.toList());
	// Create a list of local hidden variables
	List<Variable> localHidden = new ArrayList<Variable>();
	for(int i= 0; i< attrVars.size(); i++) {
		localHidden.add(vars.newGaussianVariable("LocallHidden"+i));
	}	
	// Create a global hidden variable
	Variable globalHidden = vars.newMultinomialVariable("GlobalHidden",2);
	// Create a DAG over the variables (hidden and observed)
	DAG dag = new DAG(vars);
	// Add the links
	for (int i=0; i<attrVars.size(); i++) {
		dag.getParentSet(attrVars.get(i)).addParent(globalHidden);
		dag.getParentSet(attrVars.get(i)).addParent(localHidden.get(i));
	}
	//This is needed to maintain coherence in the Model class.
	this.dag = dag;
	}
}

\end{lstlisting}

\vspace{5mm}

To learn this custom model, we simply create an object of class \texttt{CustomModel}, and invoke the method \texttt{updateModel} as shown in \code~\ref{code:learningModel}.

\begin{lstlisting}[
caption={Learning a custom model from data.}, 
label=code:learningModel,
style=java, 
]
Model model = new CustomModel(data.getAttributes());
model.updateModel(data);
\end{lstlisting}

\vspace{-2mm}

\subsection{Inference}

\pkg{AMIDST} provides inference algorithms exploiting multi-core architectures. As an example, \linebreakGen \code~\ref{code:staticInference} shows how to make inference in the Gaussian mixture \linebreakGen learnt in  \code~\ref{code:StaticModelLearning}. All the functionality for making inference is defined in the interface \texttt{eu.amidst.core.inference.InferenceAlgorithm}. Any class implementing a particular inference algorithm should implement this interface.  VMP is the algorithm used in code below. This can be changed by using another constructor in line 10 (e.g., \texttt{new HuginInference()} or \texttt{new ImportanceSampling()}). By invoking the method \texttt{setModel(BayesianNetwork model)}, we specify in which model we aim to make inference (line 11). The evidence is then set in line 12. Finally, the inference is done by invoking the method \texttt{runInference()}. Afterwards,  we can obtain the posterior distribution of any variable by calling the method \texttt{getPosterior(Variable var)}.

\begin{lstlisting}[
caption={Inference in a (static) BN using the VMP algorithm. }, 
label=code:staticInference,
style=java, 
]
	// Set the variables of interest
Variable varTarget = bn.getVariables().getVariableByName("HiddenVar");

// Set the evidence
Assignment assignment = new HashMapAssignment();
assignment.setValue(bn.getVariables().getVariableByName("GaussianVar8"), 8.0);
assignment.setValue(bn.getVariables().getVariableByName("GaussianVar9"), -1.0);

// Set the inference algorithm
InferenceAlgorithm infer = new VMP(); 
infer.setModel(bn);
infer.setEvidence(assignment);

// Run the inference
infer.runInference();
Distribution p = infer.getPosterior(varTarget);
System.out.println("P(HiddenVar|GaussianVar8=8.0, GaussianVar9=-1.0) = "+p);
\end{lstlisting}

Analogously, the inference in a dynamic model is illustrated in \code~\ref{code:dynInference}. In particular, this example shows how to use the Factored Frontier algorithm with importance sampling. One difference w.r.t. the static case is that the assignment includes information about the current time (line 16).  Inference is then done by invoking  \texttt{runInferenrece}.  When dealing with dynamic models, we might be interested in computing the posterior distributions at the current instant of time (line 23) or in the future (line 27).

\begin{lstlisting}[
caption={Inference in a dynamic model. }, 
label=code:dynInference,
style=java, 
]
// Select the inference algorithm
InferenceAlgorithmForDBN infer = 
			new FactoredFrontierForDBN(new ImportanceSampling()); 
			
infer.setModel(dbn);
// Set the variables of interest
Variable varTarget = dbn.getDynamicVariables().getVariableByName("gaussianHiddenVar1");

for(int t=0; t<10; t++) {
	// Set the evidence
	HashMapDynamicAssignment assignment = new HashMapDynamicAssignment(2);
	assignment.setValue(dbn.getDynamicVariables().getVariableByName("GaussianVar9"),
																				 rand.nextDouble());
	assignment.setValue(dbn.getDynamicVariables().getVariableByName("GaussianVar8"),
																				 rand.nextDouble());
	assignment.setTimeID(t);

	// Run the inference
	infer.addDynamicEvidence(assignment);
	infer.runInference();

	// Get the posterior at current instant of time
	Distribution posterior_t = infer.getFilteredPosterior(varTarget);
	System.out.println("t="+t+" "+posterior_t);

	// Get the posterior in the future
	Distribution posterior_t_1 = infer.getPredictivePosterior(varTarget, 1);
	System.out.println("t="+t+"+1 "+posterior_t_1);

}


\end{lstlisting}

\vspace{-5mm}
\section{Documentation, Availability, and Further Developments}\label{sec:documentation}

The \pkg{AMIDST} Toolbox is available at \url{http://www.amidsttoolbox.com} under the Apache Software License version 2.0 and the documented source code is hosted on GitHub\footnote{\url{https://github.com/amidst/toolbox}}. The website includes a tutorial explaining how to install the toolbox, and provides a large set of source code examples of the use of the toolbox API, including example implementations of well-known models (see Table \ref{tab:latent-var-models}).  The code examples shown in this paper can be found in a specific repository\footnote{\url{https://github.com/amidst/jss}}. This includes all the material for compiling and running the examples (scripts, java files, libraries, etc.). The toolbox is distributed using Maven\footnote{\url{https://maven.apache.org}}. The use of this technology simplifies the installation  making the interaction with external software transparent. In addition, public collaborations is supported and encouraged via GitHub Fork and Pull requests.  The toolbox has been uploaded to the MLOSS\footnote{\url{http://mloss.org/software/}} repository.  Additionally, the toolbox can be acessed from R by means of the package \pkg{ramidst}\footnote{\url{https://cran.r-project.org/web/packages/ramidst/}} which provides access to some key functionalities.



	\section*{Acknowledgments}
	This work was performed as part of the AMIDST project. AMIDST has received funding from the European Union's Seventh Framework Programme for research, technological development and demonstration under grant agreement no 619209.

	\bibliographystyle{plain}
	\bibliography{biblio}
\end{document}